\documentclass[fleqn,10pt]{SelfArx} 

\definecolor{color1}{RGB}{0,0,90} 
\definecolor{color2}{RGB}{0,20,20} 

\newlength{\tocsep} 
\usepackage{lipsum} 

\usepackage{tabularx, booktabs}
\newcolumntype{Y}{>{\centering\arraybackslash}X}

\usepackage[noend]{algpseudocode}
\usepackage{algorithm}

\algrenewcommand\alglinenumber[1]{{\textit{\small #1}}}

\usepackage{varwidth}
\usepackage{url}

\usepackage{tabto}

\JournalInfo{\tiny ~} 
\Archive{\tiny ~} 

\PaperTitle{Retaining Experience and Growing Solutions} 

\Authors{Robyn Ffrancon\textsuperscript{1}} 
\affiliation{\textsuperscript{1}\textit{TAO team, INRIA Saclay, Univ. Paris-Sud, France}} 
\affiliation{\textbf{Correspondence}: rffrancon@gmail.com - robynffrancon.com} 

\Keywords{Genetic Programming --- Semantic Backpropagation --- Local Search --- Meta-GP --- Neural Networks} 
\newcommand{\keywordname}{Keywords} 

\Abstract{Generally, when genetic programming (GP) is used for function synthesis any valuable experience gained by the system is lost from one problem to the next, even when the problems are closely related. With the aim of developing a system which retains beneficial experience from problem to problem, this paper introduces the novel \textit{Node-by-Node Growth Solver~(NNGS)} algorithm which features a component, called the \textit{controller}, which can be adapted and improved for use across a set of related problems. NNGS grows a single solution tree from root to leaves. Using semantic backpropagation and acting locally on each node in turn, the algorithm employs the controller to assign subsequent child nodes until a fully formed solution is generated. 

The aim of this paper is to pave a path towards the use of a neural network as the controller component and also, separately, towards the use of meta-GP as a mechanism for improving the controller component. A proof-of-concept controller is discussed which demonstrates the success and potential of the NNGS algorithm. In this case, the controller constitutes a set of hand written rules which can be used to deterministically and greedily solve standard Boolean function synthesis benchmarks. Even before employing machine learning to improve the controller, the algorithm vastly outperforms other well known recent algorithms on run times, maintains comparable solution sizes, and has a 100\% success rate on all Boolean function synthesis benchmarks tested so far.}

\begin{document}

\flushbottom 

\maketitle 

\tableofcontents 

\thispagestyle{empty} 

\section{Motivation}

Most genetic programming (GP)~\cite{classical_gp_Koza_1} systems don't adapt or improve from solving one problem to the next. Any experience which could have been gained by the system is usually completely forgotten when that same system is applied to a subsequent problem, the system effectively starts from scratch each time. 

For instance, consider the successful application of a classical GP system to a standard n-bits Boolean function synthesis benchmark (such as the 6-bits comparator as described in~\cite{bpgp_Krawiec}). The population which produced the solution tree is not useful for solving any other n-bits Boolean benchmark (such as the 6-bits multiplexer). Therefore, in general, an entirely new and different population must be generated and undergo evolution for each different problem. This occurs because the system components which adapt to solve the problem (a population of trees in the case of classical GP) become so specialised that they are not useful for any other problem.

This paper addresses this issue by introducing the \textit{Node-by-Node Growth Solver~(NNGS)} algorithm, which features a component called the \textit{controller}, that can be improved from one problem to the next within a limited class of problems.

NNGS uses \textit{Semantic Backpropagation (SB)} and the controller, to grow a single S-expression solution tree starting from the root node. Acting locally at each node, the controller makes explicit use of the target output data and input arguments data to determine the properties (i.e. operator type or argument, and semantics) of the subsequently generated child nodes.

The proof-of-concept controller discussed in this paper constitutes a set of deterministic hand written rules and has been tested, as part of the NNGS algorithm, on several popular Boolean function synthesis benchmarks. This work aims to pave the way towards the use of a neural network as an adaptable controller and/or, separately, towards the use of meta-GP for improving the controller component. In effect, the aim is to exploit the advantages of black-box machine learning techniques to generate small and examinable program solutions.

The rest of this paper will proceed as follows: Section~\ref{related_work_section} outlines other related research. Section~\ref{sb_section} details semantic backpropagation. A high level overview of the NNGS system is given in Section~\ref{nngs_section}, and Section~\ref{controller_section} describes the proof-of-concept controller. Section~\ref{experiments_section} details the experiments conducted. The experimental results and a discussion is given in Section~\ref{results_section}. Section~\ref{further_work_section} concludes with a description of potential future work.

\section{Related Work} \label{related_work_section} 

The isolation of useful subprograms/sub-functions is a related research theme in GP. However, in most studies subprograms are not reused across different problems. In~\cite{hier_auto_func_koza} for instance, the hierarchical automatic function definition technique was introduced so as to facilitate the development of useful sub-functions whilst solving a problem. Machine learning was employed in~\cite{bpgp_Krawiec} to analyse the internal behaviour (semantics) of GP programs so as to automatically isolate potentially useful problem specific subprograms.

SB was used in~\cite{sbgp_operators_Pawlak} to define intermediate subtasks for GP. Two GP search operator were introduced which semantically searched a library of subtrees which could be used to solve the subtasks. Similar work was carried out in~\cite{memetic_semantic_gp_Ffrancon, glti_Ffrancon}, however in these cases subtree libraries were smaller and static, and only a single tree was iteratively modified as opposed to a population of trees.

\section{Semantic Backpropagation (SB)} \label{sb_section}

Semantic backpropagation (SB) within the context of GP is an active research topic~\cite{sbgp_operators_Pawlak, agx_sb_Krawiec, memetic_semantic_gp_Ffrancon, glti_Ffrancon}.

Consider the \textit{output array} produced by the root node of a solution tree, where each element within the array corresponds to the output from one fitness case. This output array is the semantics of the root node. If the solution is perfectly correct, the output array will correspond exactly to the target output array of the problem at hand. In a programmatic style, the output array of a general node $node\_x$ will be denoted as $node\_x.\text{outputs}$ and the output from fitness case $i$ will be denoted by $node\_x.\text{outputs}[i]$.

Each node within a tree produces an output array, a feature which has been exploited in~\cite{bpgp_Krawiec} to isolate useful subtrees. The simplest example of a tree (beyond a single node) is a triplet of nodes: a parent node $node\_a$, the left child node $node\_b$, and the right child node $node\_c$. 

As a two fitness case example, suppose that a triplet is composed of a parent node $node\_a$ representing the operator $AND$, a left child node $node\_b$ representing input argument $A1 = [0,1]$, and a right child node $node\_c$ representing input argument $A2 = [1,1]$. The output array of the parent node is given by:

\begin{equation}
\begin{aligned}
node\_a.\text{outputs} &= node\_b.\text{outputs} \text{ }AND\text{ } node\_c.\text{outputs} \\
      &= [0,1] \text{ }AND\text{ } [1,1] \\
      &= [0,1].
\end{aligned}
\end{equation}

On the other hand, given the output array from the parent node of a triplet $node\_a$, it is possible to backpropagate the semantics so as to generate output arrays for the child nodes, if the reverse of the parent operator is defined carefully. This work will exclusively tackle function synthesis problems within the Boolean domain, and therefore, the standard~\cite{hier_auto_func_koza, bpgp_Krawiec} set of Boolean operators will be used: $AND$, $OR$, $NAND$, and $NOR$.

\begin{figure}[!ht]
\small
\centerline{
\begin{tabular}{cc}
\begin{tabularx}{0.15\textwidth}{|Y||Y|Y|}
\hline
\multicolumn{3}{|c|}{$\hat{b}, \hat{c} = AND^{-1}(\hat{a})$} \\ \hline
$\hat{a}$ & $\hat{b}$ & $\hat{c}$ \\ \hline
1 & 1 & 1 \\
0 & 0 & \# \\
0 & \# & 0 \\
\# & \# & \# \\
\hline
\end{tabularx}
&
\begin{tabularx}{0.15\textwidth}{|Y||Y|Y|}
\hline
\multicolumn{3}{|c|}{$\hat{b}, \hat{c} = OR^{-1}(\hat{a})$} \\ \hline
$\hat{a}$ & $\hat{b}$ & $\hat{c}$ \\ \hline
1 & 1 & \# \\
1 & \# & 1 \\
0 & 0 & 0 \\
\# & \# & \# \\
\hline
\end{tabularx} \\
& \\ 
\begin{tabularx}{0.16\textwidth}{|Y||Y|Y|}
\hline
\multicolumn{3}{|c|}{$\hat{b}, \hat{c} = NAND^{-1}(\hat{a})$} \\ \hline
$\hat{a}$ & $\hat{b}$ & $\hat{c}$ \\ \hline
1 & 0 & \# \\
1 & \# & 0 \\
0 & 1 & 1 \\
\# & \# & \# \\
\hline
\end{tabularx}
&
\begin{tabularx}{0.15\textwidth}{|Y||Y|Y|}
\hline
\multicolumn{3}{|c|}{$\hat{b}, \hat{c} = NOR^{-1}(\hat{a})$} \\ \hline
$\hat{a}$ & $\hat{b}$ & $\hat{c}$ \\ \hline
1 & 0 & 0 \\
0 & 1 & \# \\
0 & \# & 1 \\
\# & \# & \# \\
\hline
\end{tabularx}
\end{tabular}
}
\caption{Function tables for the reverse operators: $\text{AND}^{-1}$, $\text{OR}^{-1}$, $\text{NAND}^{-1}$, and $\text{NOR}^{-1}$.\label{reverse_ops_fig}}
\end{figure}

Figure~\ref{reverse_ops_fig} gives function tables for the element-wise reverse operators: $AND^{-1}$, $OR^{-1}$, $NAND^{-1}$, and $NOR^{-1}$ (their use with 1D arrays as input arguments follows as expected). As an example use of these operators the previous example will be worked in reverse: given the output array of $node\_a$, the arrays $node\_b.\text{outputs}$ and $node\_c.\text{outputs}$ are calculated as:

\begin{equation}
\begin{aligned}
node\_b.\text{outputs}\text{,~}node\_c.\text{outputs} &= AND^{-1}(node\_a.\text{outputs}) \\
&= AND^{-1}([0,1]) \\
&= [0, 1] \text{, } [\#, 1] \\
&\text{or} \\
&= [\#, 1] \text{, } [0, 1]. \\
\end{aligned}
\end{equation}

The hash symbol $\#$ in this case indicates that either $1$ or $0$ will do. Note that two different possible values for $node\_b.\text{outputs}$ and $node\_c.\text{outputs}$ exist because $AND^{-1}(0) = (0, \#) \text{ or } (\#, 0)$. This feature occurs as a result of rows 4 and 5 of the $NAND^{-1}$ function table. Note that each of the other reverse operators have similar features, specifically for: $OR^{-1}(1)$, $NAND^{-1}(1)$, and $NOR^{-1}(0)$. 

Note also, that for any array loci $i$ in $node\_a.\text{outputs}$ where $node\_a.\text{outputs}[i] = \#$, it is true that $node\_b.\text{outputs}[i] = \#$ and $node\_c.\text{outputs}[i] = \#$. For example, $NOR^{-1}([1, \#]) = ([0, \#], [0, \#])$.

Using the reverse operators in this way, output arrays can be assigned to the child nodes of any parent node. The child output arrays will depend on two decisions: Firstly, on the operator assigned to the parent node, as this is the operator that is reversed. And secondly, on the choices made (note the $AND^{-1}(0)$ example above), at each loci, as to which of the two child output arrays contains the $\#$ value. These decisions are made by the controller component.

Using these reverse operators for SB can only ever produce a pair of output arrays which are different from the problem target outputs in two ways. Firstly, the output arrays can be a flipped (using the $NOT$ gate on each bit) or an un-flipped versions of the problem target outputs. Secondly, some elements of the output arrays will be $\#$ elements.

\section{Node-by-Node Growth Solver (NNGS)} \label{nngs_section}

\begin{figure}[!ht]
 \includegraphics[width=1\columnwidth]{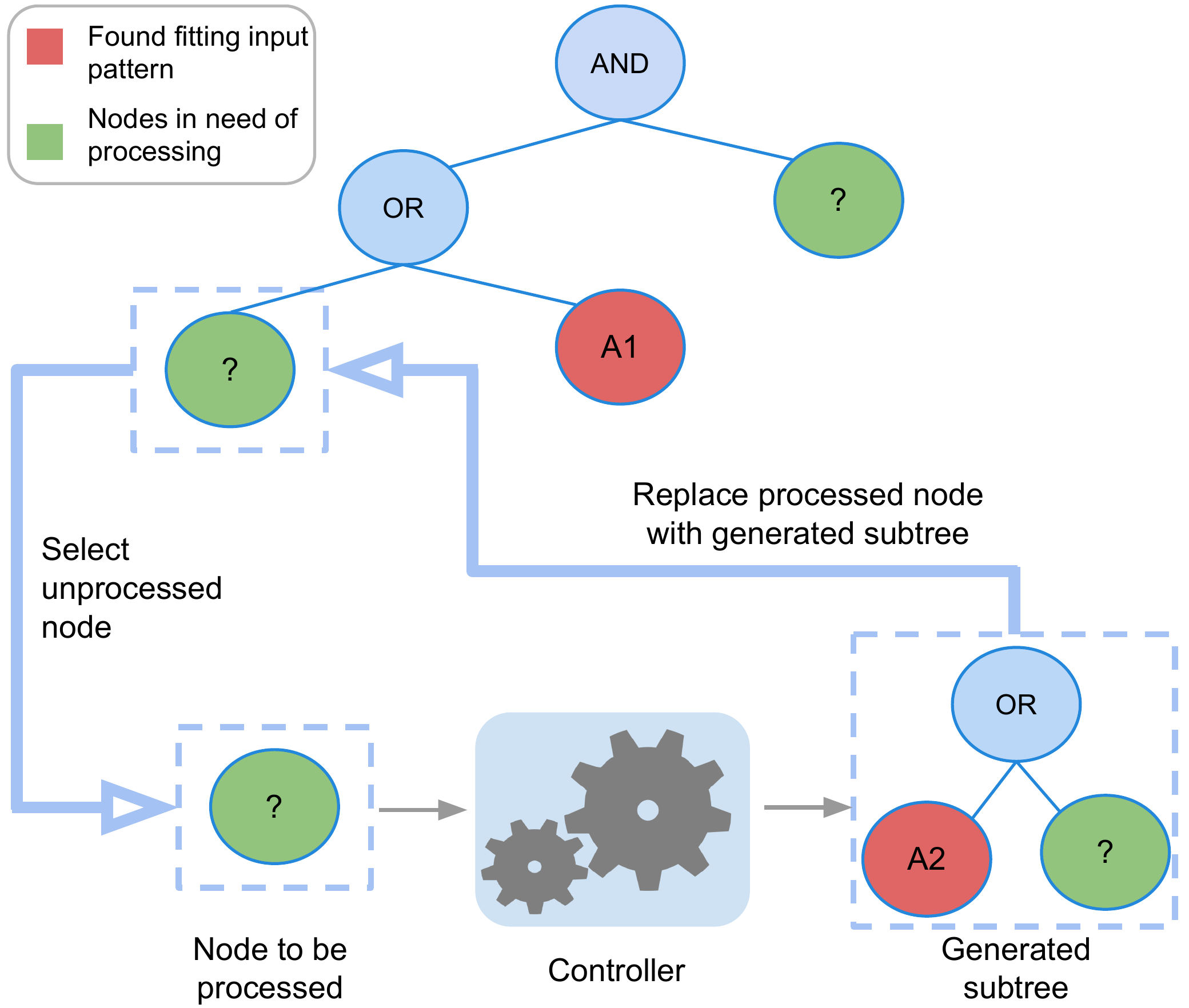}
\caption{A visual representation of the NNGS algorithm during the development of a solution tree.}
\label{nngs_visual_fig}
\end{figure}

A visual representation of the NNGS algorithm can be seen in Fig.~\ref{nngs_visual_fig}, which shows a snapshot of a partially generated solution tree. This tree, in it's unfinished state, is composed of: $AND$ and $OR$ operators, an input argument labelled $A1$, and two unprocessed nodes. The basic role of the NNGS algorithm is to manage growing the solution tree by passing unprocessed nodes to the controller and substituting back the generated/returned node triplet. 

\begin{algorithm}[!ht]
\caption{The Node-by-Node Growth Solver \newline
NNGS(target\_outputs, controller)}
\label{nngs_alg}
\begin{algorithmic}[1]

\State{solution\_tree $\gets$ \{\}}
\State{$root\_node$.outputs $\gets$ target\_outputs} \label{setting_root_node_alg}
\State{unprocessed\_nodes $\gets$ \{$root\_node$\}}
\item[]

\While{len(unprocessed\_nodes) $>$ 0} 
	\State{$node\_a \gets$ unprocessed\_nodes.pop()}
    \item[] \Comment{check for leaf node}
    
    \If{$node\_a$.type = 'argument'}  \label{avoid_arg_nodes_alg}
    	\State{solution\_tree.insert($node\_a$)}
        \State{\textbf{continue}} \Comment{move on to next node}
    \EndIf
    \item[]
    
    \State \begin{varwidth}[t]{\linewidth}
      $node\_a$, $node\_b$, $node\_c \gets$ \par
        \hskip\algorithmicindent controller($node\_a$, target\_outputs) \label{controller_alg}
      \end{varwidth}
    \item[]
    
    \State{unprocessed\_nodes.insert(\{$node\_b$, $node\_c$\})}
    \State{solution\_tree.insert($node\_a$)}
    \item[]
    
\EndWhile

\State{\Return solution\_tree}

\end{algorithmic}
\end{algorithm}

Algorithm~\ref{nngs_alg} gives a simple and more thorough explanation of NNGS. In line~\ref{setting_root_node_alg} the output values of the solution tree root node are set to the target output values of the problem at hand. The output values of a node are used, along with the reverse operators, by the controller (line~\ref{controller_alg}) to generate the output values of the returned child nodes. The controller also sets the node type (they are either operators or input arguments) of the input parent node and generated child nodes.

Nodes which have been defined by the controller as input arguments (with labels: $A1$, $A2$, $A3$... etc.) can not have child nodes (they are by definition leaf nodes) and are therefore not processed further by the controller (line~\ref{avoid_arg_nodes_alg}). When every branch of the tree ends in an input argument node, the algorithm halts.

Note that the controller may well generate a triplet where one or more of the child nodes require further processing. In this case the NNGS algorithm will pass these nodes back to the controller at a later stage before the algorithm ends. In effect, by using the controller component the NNGS algorithm simply writes out the solution tree.

\section{Proof-Of-Concept Controller} \label{controller_section}

Given an unprocessed node, the controller generates two child nodes and their output arrays using one of the four reverse operators. It also sets the operator type of the parent node to correspond with the chosen reverse operator that is used. 

The ultimate goal of the controller is to assign an input argument to each generated child node. For example, suppose that the controller generates a child node with an output array $node\_b.\text{outputs}=[0,1,1,\#]$ and that an input argument is given by $A1=[0,1,1,0]$. In this case, $node\_b$ can be assigned (can represent) the input argument $A1$ because $[0,1,1,\#]=[0,1,1,0]$. The algorithm halts once each leaf node has been assigned an input argument.

Before giving a detailed description of the proof-of-concept controller, there are a few important general points to stress: Firstly, the entire decision making process is deterministic. Secondly, the decision making process is greedy (the perceived best move is taken at each opportunity). Thirdly, the controller does not know the location of the input node within the solution tree. The controller has priori knowledge of the input argument arrays, the operators, and the reverse operators only. Furthermore, the controller, in its current state, does not memorise the results of it's past decision making. In this regard, when processing a node, the controller has knowledge of that node's output array only. In this way, the controller acts locally on each node. Multiple instances of the controller could act in parallel by processing all unprocessed nodes simultaneously.



\subsection{Step-by-step}

\begin{figure}[!ht]
 \includegraphics[width=1\columnwidth]{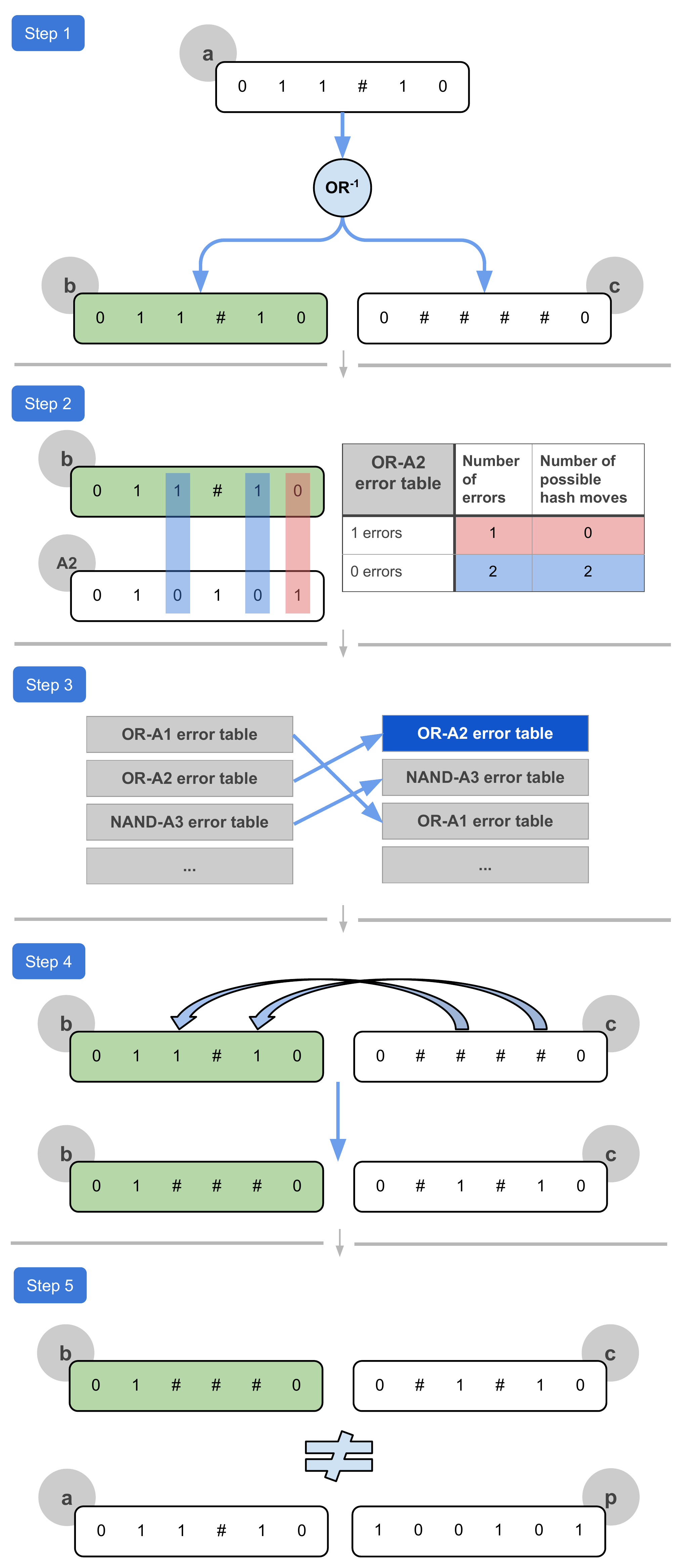}
\caption{Diagrammatic aid for the proof-of-concept controller.}
\label{controller_visual_fig}
\end{figure}

This subsection will give a step-by-step run-through of the procedure undertaken by the proof-of-concept controller. Figure~\ref{controller_visual_fig} serves as a diagrammatic aid for each major step.

\subsubsection{Step 1}

Given an input (parent) node $node\_a$, and for each reverse operator in Table~\ref{reverse_ops_fig}, the first step taken by the controller is to generate output arrays for each of the child nodes. In the example given in step 1 of Fig.~\ref{controller_visual_fig} only the $OR^{-1}$ reverse operator is used. The $OR^{-1}$ reverse operator generates $\#$ values in the child output arrays due to the following property $OR^{-1}(1) = (1, \#) \text{ or } (\#, 1)$. In this step, whenever the opportunity arises (regardless of the reverse operator employed), all generated $\#$ values within the child output arrays will be placed in the output array of the right child node $node\_c$. For example in the case of $OR^{-1}(1)$: $(1, \#)$ will be used and not $(\#, 1)$.

Note that the reverse operators propagate all $\#$ elements from parent to child nodes. This feature is exemplified in step 1 of Fig.~\ref{controller_visual_fig} by the propagation of the $\#$ value at locus 4 of $node\_a.\text{outputs}$ to loci 4 of both $node\_b.\text{outputs}$ and $node\_c.\text{outputs}$. 

\subsubsection{Step 2}

By this step, the controller has generated four different (in general) $node\_b.\text{outputs}$ arrays, one for each reverse operator. The goal for this step is to compare each of those arrays to each of the possible input argument arrays ($A1$, $A2$... etc). As an example, in step 2 of Fig.~\ref{controller_visual_fig} the generated $node\_b.\text{outputs}$ array is compared to the $A2$ input argument array. 

Two separate counts are made, one for the number of erroneous 0 argument values $E_0$ and one for the number of erroneous 1 argument values $E_1$ (highlighted in blue and red respectively in Fig.~\ref{controller_visual_fig}). Two further counts are made of the number of erroneous $node\_b$ loci, for 0 and 1 input arguments values, which could have been $\#$ values (and therefore not erroneous) had the controller not specified in step 1 that all $\#$ values should be placed in the $node\_c.\text{outputs}$ array whenever possible. These last two statistics will be denoted by $M_0$ and $M_1$ for 0 and 1 input arguments values respectively. These four statistics form an \textit{error table} for each reverse operator-input argument pair.

\subsubsection{Step 3}

In this step, the controller sorts the error tables by a number of statistics. Note that $M_0 - E_0$ and $M_1 - E_1$ are the number of remaining erroneous 0 argument values and erroneous 1 argument values respectively if all $\#$ values were moved from the $node\_c.\text{outputs}$ array to the $node\_b.\text{outputs}$ array whenever possible. To simplify matters we note that

\begin{equation}
\begin{aligned}
&\text{if~~~} M_1 - E_1 \le M_0 - E_0 \\
&\text{~~~~~let } k=1 \text{, } j=0  \\
&\text{otherwise} \\
&\text{~~~~~let } k=0 \text{, } j=1.  \\
\end{aligned}
\end{equation}

Each error table is ranked by (in order, all increasing): $M_k - E_k$, $E_k$, $M_j - E_j$, $E_j$, and the number of $\#$ values in $node\_c.\text{outputs}$. In a greedy fashion, the very best error table (lowest ranked) will be select for the next step (in Fig.~\ref{controller_visual_fig} the $OR\text{-}A2$ error table is selected). Note that the ranked list of error tables might need to be revisited later from step 5.

\subsubsection{Step 4}

The error table selected in step 3 effectively serves as an instruction which details how the $node\_b.\text{outputs}$ and \\$node\_c.\text{outputs}$ arrays should be modified. The goal of the controller is to move the minimum number of $\#$ values from the $node\_c.\text{outputs}$ array to the $node\_b.\text{outputs}$ array such as to satisfy the error count for either $1$'s or $0$'s in one of the input arguments. In the example given in Fig.~\ref{controller_visual_fig}, two $\#$ values in $node\_c.\text{outputs}$ are swapped with 1's in $node\_b.\text{outputs}$.

\subsubsection{Step 5}

In this step, the algorithm checks that the generated \\$node\_b.\text{outputs}$ and $node\_c.\text{outputs}$ arrays do not exactly equal either the parent node $node\_a$ or the grand parent node $node\_p$ (if it exists). If this check fails, the algorithm reverts back to step 3 and chooses the next best error table.

\subsubsection{Step 6}

The final step of the algorithm is to appropriately set the operator type of $node\_a$ given the final reverse operator that was used. In this step the algorithm also checks whether either (or both) of the child nodes can represent input arguments given their generated output arrays.

\section{Experiments} \label{experiments_section}

The Boolean function synthesis benchmarks solved in this paper are standard benchmarks within GP research~\cite{hier_auto_func_koza, bpgp_Krawiec, sbgp_operators_Pawlak, memetic_semantic_gp_Ffrancon}. They are namely: the comparator 6bits and 8bits (CmpXX), majority 6bits and 8bits (MajXX), multiplexer 6bits and 11bits (MuxXX), and even-parity 6bits, 8bit, 9bits, and 10bits (ParXX).
\newline\newline
\noindent Their definitions are succinctly given in~\cite{bpgp_Krawiec}:

``For an v-bit comparator Cmp v, a program is required to return true if the v/2 least significant input bits encode a number that is smaller than the number represented by the v/2 most significant bits. In case of the majority Maj v problems, true should be returned if more that half of the input variables are true. For the multiplexer Mul v, the state of the addressed input should be returned (6-bit multiplexer uses two inputs to address the remaining four inputs, 11-bit multiplexer uses three inputs to address the remaining eight inputs). In the parity Par v problems, true should be returned only for an odd number of true inputs."

The even-parity benchmark is often reported as the most difficult benchmark~\cite{hier_auto_func_koza}.

\section{Results and Discussion} \label{results_section}

\setlength{\tabcolsep}{4pt}
\begin{table*}[tb!]
\centering
\caption{Results for the NNGS algorithm when tested on the Boolean benchmarks, perfect solution were obtained for each run. BP4A columns are the results of the best performing algorithm from~\cite{bpgp_Krawiec} (* indicates that not all runs found perfect solution). The RDO$_p$ column is taken from the best performing (in terms of fitness) scheme in~\cite{sbgp_operators_Pawlak} (note that in this case, average success rates and average run times were not given).}
\label{results_table}
\small
\begin{tabular}{c|c|c|c||c|c|c|c|}
\cline{2-8}
\multicolumn{1}{l|}{} & \multicolumn{3}{c||}{Run time {[}seconds{]}} & \multicolumn{4}{c|}{Program size {[}nodes{]}} \\
& \textbf{NNGS} & BP4A & ILTI & \textbf{NNGS} & BP4A & ILTI & RDO \\ \hline
\multicolumn{1}{|c|}{Cmp06} & 0.06  & 15   & 9   & 99   & 156 & 59  & 185 \\ 
\multicolumn{1}{|c|}{Cmp08} & 0.86  & 220  & 20  & 465  & 242 & 140 & 538 \\ \hline
\multicolumn{1}{|c|}{Maj06} & 0.19  & 36   & 10  & 271  & 280 & 71  & 123 \\ 
\multicolumn{1}{|c|}{Maj08} & 3.09  & 2019*& 27  & 1391 & 563*& 236 & -   \\ \hline
\multicolumn{1}{|c|}{Mux06} & 0.21  & 10   & 9   & 333  & 117 & 47  & 215 \\ 
\multicolumn{1}{|c|}{Mux11} & 226.98& 9780 & 100 & 12373& 303 & 153 & 3063\\ \hline
\multicolumn{1}{|c|}{Par06} & 0.35  & 233  & 17  & 515  & 356 & 435 & 1601\\ 
\multicolumn{1}{|c|}{Par08} & 5.73  & 3792*& 622 & 2593 & 581*& 1972& -   \\ 
\multicolumn{1}{|c|}{Par09} & 25.11 & -    & 5850& 5709 & -   & 4066& -   \\ 
\multicolumn{1}{|c|}{Par10} & 120.56& -    & -   & 12447& -   & -   & -   \\ \hline
\end{tabular}
\end{table*}

The results are given in Table~\ref{results_table} and show that the NNGS algorithm finds solutions quicker than all other algorithms on all benchmarks with the exception of the ILTI algorithm on the Mux11 benchmark. A significant improvement in run time was found for the Par08 benchmark. 

The solution sizes produced by the NNGS algorithm are always larger than those found by the BP4A and ILTI algorithms with the exception of the Cmp06 results. The RDO scheme and ILTI algorithm both relay on traversing large tree libraries which make dealing with large bit problems very computationally intensive. As such, these methods do not scale well in comparison to the NNGS algorithm. 

It is a clear that NNGS is weakest on the Mux11 benchmark. In this case a very large solution tree was found which consisted of 12,373 nodes. The multiplexer benchmark is significantly different from the other benchmarks by the fact that only four input arguments are significant to any single fitness case: the three addressing bits and the addressed bit. Perhaps this was the reason why the chosen methodology implemented in the controller resulted with poor results in this case.

\section{Further Work} \label{further_work_section}

There are two possible branches of future research which stem from this work, the first centres around meta-GP. As a deterministic set of rules, the proof-of-concept controller is eminently suited to be encoded and evolved as part of a meta-GP system. The difficulty in this case will be in appropriately choosing the set of operators which would be made available to the population of controller programs.

The second avenue of research which stems from this work involves encoding the current proof-of-concept controller within the weights of a neural network (NN). This can be achieved through supervised learning in the first instance by producing training sets in the form of node triplets using the current controller. A training set would consist of randomly generated output arrays and the proof-of-concept controller generated child output arrays. In this way, the actual Boolean problem solutions do not need to be found before training.

As part of the task of find a better controller, the weights of the NN could be evolved using genetic algorithms (GA), similar to the method employed by~\cite{evolve_nn_Koutnik}. The fitness of a NN weight set would correspond to the solution sizes obtained by the NNGS algorithm when employing the NN as a controller: the smaller the solutions, the better the weight set fitness. Using the proof-of-concept controller in this way would ensure that the GA population would have a reasonably working individual within the initial population.

A NN may also serve as a reasonable candidate controller for extending the NNGS algorithm to continuous symbolic regression problems. In this case, the input arguments of the problem would also form part of the NN's input pattern.

\section*{Acknowledgments}
I wish to thank Marc Schoenauer for interesting discussions and excellent supervision during my masters internship at INRIA, Paris, France where this work was conducted.

\bibliographystyle{plain}

\end{document}